# Semi-Supervised Anomaly Detection in Brain MRI Using a Domain-Agnostic Deep Reinforcement Learning Approach

Zeduo Zhang[1,2], Yalda Mohsenzadeh[1,2,*]

[1]Department of Computer Science, Western University, London, Ontario, Canada, [2]Vector Institute for Artificial Intelligence, Toronto, Ontario, Canada

*corresponding author: ymohsenz@uwo.ca

**Article Type:** Original Research

## Summary Statement

A domain-agnostic semi-supervised anomaly detection framework using deep reinforcement learning achieves robust performance on brain MRI and generalizes well to industrial data.

## Key Points

- We propose a semi-supervised approach that integrates Deep Q-Networks with feature representation for surface anomaly detection, improving detection accuracy on brain MRI volumes.

- Our method demonstrates strong generalization across domains, achieving 88.7% pixel-level and 96.7% image-level AUROC on brain MRI datasets, and 99.8% and 99.3% respectively on industrial datasets - outperforming prior state-of-the-art unsupervised and semi-supervised approaches.

- The approach effectively addresses challenges posed by MRI data and limited annotations, and includes sufficient ablation studies to demonstrate its efficiency, robustness and generalizability.

## Abbreviations

AD = Anomaly Detection, DRL = Deep Reinforcement Learning, DQN = Deep Q-Network, T1 = T1-weighted Imaging, T2 = T2-weighted Imaging, SOTA = State of The Art,

**Semi-Supervised Anomaly Detection in Brain MRI Using a Domain-Agnostic Deep Reinforcement Learning Approach.**

## Abstract

**Purpose:** To develop a domain-agnostic, semi-supervised anomaly detection framework that integrates deep reinforcement learning (DRL) to address challenges such as large-scale data, overfitting, and class imbalance, focusing on brain MRI volumes.

**Material and Methods:** This retrospective study used publicly available brain MRI datasets collected between 2005 and 2021. The IXI dataset provided 581 T1-weighted and 578 T2-weighted MRI volumes (from healthy subjects) for training, while the BraTS 2021 dataset provided 251 volumes for validation and 1000 for testing (unhealthy subjects with Glioblastomas). Preprocessing included normalization, skull-stripping, and co-registering to a uniform voxel size. Experiments were conducted on both T1- and T2-weighted modalities. Additional experiments and ablation analyses were also carried out on the industrial datasets. The proposed method integrates DRL with feature representations to handle label scarcity, large-scale data and overfitting. Statistical analysis was based on several detection and segmentation metrics including AUROC and Dice score.

**Results:** The proposed method achieved an AUROC of 88.7% (pixel-level) and 96.7% (image-level) on brain

MRI datasets, outperforming State-of-The-Art (SOTA) methods. On industrial surface datasets, the model also showed competitive performance (AUROC = 99.8% pixel-level, 99.3% image-level) on MVTec AD dataset, indicating strong cross-domain generalization. Studies on anomaly sample size showed a monotonic increase in AUROC as more anomalies were seen, without evidence of overfitting or additional computational cost.

**Conclusion:** The domain-agnostic semi-supervised approach using DRL shows significant promise for MRI anomaly detection, achieving strong performance on both medical and industrial datasets. Its robustness, generalizability and efficiency highlight its potential for real-world clinical applications.

## Introduction

Anomaly Detection (AD) plays an increasingly important role in medical imaging, particularly in brain MRI, where anomalies often manifest as subtle changes in shape, texture, or structure. These abnormalities can represent early signs of disease but are often difficult to detect and label due to their irregular appearance and the high cost of expert annotation. As a result, traditional supervised methods that rely on fully labeled datasets are often impractical in clinical settings.

Unsupervised and semi-supervised approaches have emerged as promising alternatives, detecting deviations from normal patterns with little or no labeled anomaly data. However, traditional unsupervised methods [1,2] often fail to incorporate prior medical knowledge and may misclassify imaging artifacts or benign variations as pathological findings. This limitation is especially problematic in critical healthcare scenarios where distinguishing clinically relevant anomalies is vital. Semi-supervised methods [3,4] mitigate this issue by using a small set of labeled anomalies, but they frequently struggle with overfitting problem, limiting their generalizability to unseen anomalies across institutions and patient populations.

To address these limitations, we introduce a DRL-based semi-supervised anomaly detection framework. DRL [5,6] offers a fundamentally different approach by learning optimal detection strategies through sequential interaction with the data environment, optimizing a reward signal rather than relying on static labels. Through adaptive interaction with the data, our algorithm can focus on rare or uncertain cases, improving robustness, reducing overfitting and addressing imbalance. This enables more effective anomaly identification even when labeled examples are scarce.

In the context of brain MRI, our method leverages representation learning and DRL-driven policy learning to detect subtle and clinically meaningful anomalies. It does so while maintaining strong generalization across datasets, avoiding reliance on any specific modality or disease type. Moreover, our framework is domain-agnostic and scalable, showing promise not only for neuroimaging but also for broader anomaly detection tasks in radiology and beyond.

Therefore, our purpose is to develop a domain-agnostic, semi-supervised anomaly detection framework that integrates deep reinforcement learning to address challenges such as large-scale data, overfitting, and class imbalance, focusing on brain MRI volumes.

## **Materials and methods**

**Datasets and Preprocessing**

We evaluated our method on both industrial and medical datasets to assess its generalizability. To demonstrate the cross-domain generalizability of our novel anomaly detection method, we also evaluated it on two benchmark industrial datasets. For industrial anomaly detection, we used MVTec AD and BTAD. MVTec AD comprises 15 subsets across object and texture categories, with pixel-wise annotations for detective regions. BTAD includes three product classes with body and surface anomalies

For the medical domain, we trained on T1- and T2-weighted axial slices from the IXI dataset (healthy subjects) and tested on abnormal slices from the BraTS 2021 dataset. Skull stripping was performed using HD-BET tool [7] and spatial alignment to the Montreal Neurological Institute template was done using the FSL FLIRT [8,9]. After co-registration, volumes measure $182 \times 218 \times 182$ voxels. To focus on diagnostically relevant regions, we discard the first 20 and last 42 axial slices of each volume, then crop and zero-pad the remainder to a uniform size of $192 \times 192 \times 96$. After preprocessing, we obtained 55,776 and 55,488 T1 and T2 slices for training, 24,096 for validation, and 96,000 for testing. To simulate a semi-supervised setting, a small number of labeled anomalies were sampled from the validation set for training and excluded from evaluation.

The datasets used in this study, IXI and BraTS 2021, are publicly available and have been fully de-identified by the original data providers prior to release. This study used fully de-identified, publicly available imaging datasets (IXI and BraTS 2021). No identifiable patient information was accessed, and no interaction with human subjects occurred. Informed consent was not required, and the study qualifies for IRB exemption. All available volumes from both datasets were used without applying additional eligibility criteria. All included volumes had complete imaging modalities; no missing data handling was necessary. Ground truth annotations were provided by the IXI and BraTS 2021 datasets. IXI comprises healthy control subjects, and all volumes are treated as normal. BraTS 2021 includes expert-verified voxel-level tumor segmentations. No additional annotations were performed. The IXI dataset contributed 581 T1-weighted and 578 T2-weighted MRI volumes from healthy subjects, which were used exclusively for training. The BraTS 2021 dataset provided 251 annotated volumes for validation and 1,000 volumes for testing, comprising patients with glioblastomas. No sample size calculation was performed, as all available data were used. Dataset partitions were disjoint at the patient level.

**Evaluation Metrics**

We assessed detection and localization performance using image- and pixel-level that are critical to evaluate anomaly detection methods and medical applications. We include image-level AUROC (I-AUROC), image-level AUPRC (I-AUPRC), maximum DICE score ($[DICE]$) for detection performance and pixel-level AUROC (P-AUROC), pixel-level AUPRC (P-AUPRC), pixel-level $[DICE]$ for localization performance and Specificity and Sensitivity which are particularly critical in medical applications. High sensitivity minimizes false negatives (missed diagnoses), while high specificity limits false positives (unnecessary interventions), ensuring patient safety and efficient use of clinical resources.

**Problem Setup**

In AD, we start with a large collection of unlabeled data $D^u$ of predominantly normal samples (pixel-level annotation being prohibitively costly), which we refer to as the normal set in this paper. In the semi-supervised scenario, we additionally introduce a small labeled dataset $D^a$, which contains a few annotated anomalous samples with pixel-level masks. To localize surface anomalies, we train a model at the patch level to classify whether a given region is normal or abnormal. In semi-supervised setting, we usually assume $|D^a| \in \{1,10\}$ while $|D^u| \gg |D^a|$, and anomalous voxels make up a vanishingly small fraction of all voxels - creating an extreme imbalance that far exceeds that of conventional imbalanced-class classification.

**Proposed Method**

In Deep Q-Network (DQN), an agent learns to maximize cumulative rewards by interacting with an environment to determine optimal actions. The agent observes the current state and performs an action, receiving feedback in the form of a reward and the next state from the environment. This feedback is used to refine the agent's policy through exploration and exploitation.

In our framework, we treat the feature representation of each pixel or patch in an image as a state and train a neural network to classify these states as normal or anomalous. We design an environment that enhances the exploration and exploitation on image features $s_t \in \{S^a, S^u\}$, with the network receiving rewards $r_t$ based on its predictions $a_t = \arg\max_{a_t} Q(s_t, a_t; \theta)$. This reward mechanism, integrated through DQN loss $L(\theta)$, encourages accurate feature classification. The pseudocode algorithm and flow diagram are presented in Figure 1 and Figure 2

**Feature Representation (State)**

To perform anomaly localization on images using DQN, we define a "state" as the feature representation of a patch instead of the whole image. We use a locally-aware feature extractor, as described by [1,10], which leverages intermediate layers of pre-trained ImageNet model to extract meaningful path features that also encode neighborhood context rather than isolated patch information.

Each image from these datasets is denoted as $x_i \in D^a \cup D^u$, where $x_i \in R^{3 \times H \times W}$. Each sample $x_i$ is processed through the ImageNet model, and the output form the $l$-th layer, where $l \in L$, is represented as $x^{l,i} \in R^{C_l \times H_l \times W_l}$ and $(C_l, H_l, W_l)$ represents the dimensions of the features map from layer $l$. Subsequently, we aggregate this information using a patch of size $p$ to obtain a local feature vector $z^{l,i}$ that incorporates information from neighboring patches. Next, we *upscale* all feature maps with different resolutions from layers to a uniform resolution $(H_0, W_0)$ and *concatenate* them together. Thus, the extracted feature map of each sample $f^i \in R^{H_0, W_0, C}$ is defined as

$$f^i = concat\left(unscaling(z^{l,i}, (H_0, W_0) | l \in L)\right),$$

Where $C$ is the target dimension after aggregation. The state at any location $(h, w)$ on the feature map $f^i$ is denoted as $f^i_{h,w} \in R^C$. The extracted feature map is at low resolution. Since the anomalous regions are small in some datasets, instead of downscaling the ground-truth map to low resolution, we upscale the feature map to the same resolution as the original image. Then each feature $f^i_{h,w}$ has their corresponding ground truth label $gt(f^i_{h,w}) \in \{0,1\}$ where 0 for normal and 1 for abnormal. Each $f^i_{h,w}$ represents a state in DQN, so to stay aligned with DQN definition, $f^i_{h,w}$ is represented by $s_t$ in the following section. This feature extraction process is also defined as $f^i = F(x_i)$.

**Anomaly Detection Model (agent)**

The goal of our agent $A$ is to acquire the best action-value function for AD. Following [11,12], the value function can be roughly expressed as follows:

$$Q(s, a) = \max_{\pi} E[r_t + \gamma r_{t+1} + \gamma^2 r_{t+2} + \cdots \mid s_t = s, a_t = a, \pi]$$

This function, given a state $s$ and action $a \in \{a^0, a^1\}$, under a behavior policy $\pi = P(a \mid s)$, calculates the maximum expected return. The return is a cumulative sum of discounted rewards $r_t$ at step $t$, with $\gamma$ as the discount factor. This work makes use of the well-known Deep Q-Network (DQN) [12], which uses $\theta$ parameters and deep neural networks as the function approximator: $Q(s, a; \theta) = Q^*(s, a)$. It then minimizes the following loss iteratively to learn the parameters $\theta$:

$$L(\theta) = E_{(s', a', r', s'_{t+1}) \sim B}\left[r' + \gamma \max_{a'_{t+1}} \hat{Q}(s'_{t+1}, a'_{t+1}; \theta^-) - Q(s', a'; \theta)\right]$$

Where $Q$ is the action-value network with parameters $\theta$, and $\hat{Q}$ is the target network with parameters $\theta^-$ and architecture exactly same as $Q$. $\theta^-$ is updated with $\theta$ every $K$ steps. $B$ is an experience replay buffer storing past experiences, which each element stored as $e_t = (s_t, a_t, r_t, s_{t+1})$; the loss is computed using mini-batch samples drawn uniformly at random from the stored experience.

**Self-designed Environment**

DQN requires an environment to give the reward and generate next state to agent providing its state and corresponding action. Our environment is inspired by [6] with modifications for our own purposes.

**Exploration and Exploitation in State Generation:** The sampling function in our model is divided into two parts: exploitation $g_a$ and exploration $g_u$. With probability $\beta$, the environment executes $g_a$, randomly selecting $s_{t+1}$ from $D^a$. Conversely, with probability $1 - \beta$, it performs $g_u$, drawing $s_{t+1}$ from $D_u$ to identify challenging points. The function $g_u$ is defined as follows:

$$g_u(s_{t+1}|s_t, a_t, \theta^e) = \begin{cases} \arg\min_{s \in S^u} d(s_t, s; \theta^e) \text{ if } a_t = a^1 \\ \arg\max_{s \in S^u} d(s_t, s; \theta^e) \text{ if } a_t = a^0 \end{cases},$$

where $S^u$ is a subset of normal features, $\theta^e$ represents the parameters of the feature embedding function derived from the hidden layer of $Q$, and $d$ is the Euclidean distance. This function measures the distance between the feature embeddings of $s_t$ and $s$, as output by the last hidden layer of the DQN, allowing us to gauge the distance as perceived by the agent in its representation space.

Specifically, when the agent accurately predicts a normal state, $g_u$ identifies the farthest neighbor of $s_t$, enabling the exploration of potential unlearned points. Conversely, if the prediction is incorrect, $g_u$ finds the nearest neighbor of $s_t$, encouraging the agent to focus on these similar challenging points for deeper learning. In this study, we set $\beta = 0.5$ to ensure effective utilization of bot labeled anomalies and normal data. This approach is particularly beneficial for handling imbalanced datasets, as it ensures equal exploration on normal and abnormal data. Consequently, common and straightforward elements within normal data, such as background features, are not repeatedly analyzed, allowing the agent to concentrate on distinguishing more subtle features.

To enable the agent to explore large, complex datasets, we propose sampling functions $g_a$ and $g_u$ on periodically updated data subsets. Specifically, we select an anomalous image $I^a \in D^a$, and extract anomalous and $S^a$. For large normal datasets, a subset of $N$ images is sampled, yielding $S^u$ features. In addition, finding the shortest and farthest neighbors in the entire unknown dataset may cause the agent to get stuck in a specific region.

**Rewards:** We define a reward $r_t$ for our agent based on its output:

$$r_t = \begin{cases} 1 & if\ a_t = a^1\ and\ gt(s_t) = 1 \\ -1 & if\ a_t = a^0\ and\ gt(s_t) = 1 \\ -2 & if\ a_t = a^1\ and\ gt(s_t) = 0 \\ 0 & if\ a_t = a^0\ and\ gt(s_t) = 0. \end{cases}$$

This reward definition is similar to one used in [5,6]. The reward structure grants a positive reward only when the agent correctly identifies anomalies $a^1\ and\ gt(s_t) = 1$. No reward or penalty is assigned for correctly identifying normal instances $a^0\ and\ gt(s_t) = 0$, as these can be considered as common scene in environment. Incorrect predictions, whether normal or anomalous, incur a penalty. The severity of the penalty is contingent upon the specific tolerance levels for false positives and false negatives, which vary according to the scenario. In our experiments, penalties are adjusted primarily to discourage false positives.

**Anomaly Score**

In Deep Q-learning, the optimal policy is derived from the action-value function $Q^*(s, a)$ as follow:

$$\pi^*(s) = \arg\max_a Q^*(s, a).$$

In our algorithm, the optimal action $a^*$ corresponds to the class $y$ with the highest Q-value. Our reward function is designed to ensure that Q-values for correct class is higher than for any incorrect class: $Q^*(s, y) > Q^*(s, a), \forall\, a \neq y$. Thus, the learned policy satisfies: $\pi^*(s) = y$, which aligns with the classification criterion that assigns the correct label $y$ to each input state $s$.

In anomaly detection task, let $s^n$ represent normal features and $s^a$ represent abnormal features. The reward function is further designed to ensure that $Q^*(s^a, a^1) > Q^*(s^n, a^1)$. As a result, $Q^*(s, a^1)$ can be interpreted as an anomaly score.

**Code Availability**

The implementation of our method is available at https://anonymous.4open.science/r/DQL_AD-D4D0

## Results

We benchmark our approach against State-of-The-Art (SOTA) unsupervised and semi-supervised methods, including reconstruction-based (EDC [13], DAE [14]), embedding-based (PatchCore [10], SimpleNet [1], AE-FLOW [15]), knowledge-distillation (RD++ [2]) and 3D embedding-based (SimpleSliceNet [16]) methods. We also compare with semi-supervised methods like BGAD [17], DevNet [18], DRA [3], and PRN [4].

For medical datasets, we compare with methods shown effective in brain MRI (e.g., EDC, DAE, AE-FLOW). For PRN, we use only the MVTec results reported in the original paper—no further comparisons are feasible because the code isn't publicly available. SimpleNet and PatchCore are included for sharing the same feature extraction protocol. As shown in Figure 6 of the main text and Figure S2 of the Supplementary Material, we employ the base framework (DQN_Base) for experiments on the BraTS 2021 dataset, and the full variant (DQN_All) for the industrial datasets. Because T2-weighted images provide markedly better tumor contrast than T1, all analyses and ablation studies on BraTS 2021 refer to the T2 modality; T1 results are presented for completeness, but unless otherwise specified, any discussion of BraTS 2021 metrics pertains to T2. To ensure result stability and statistical reliability, we report the mean and standard deviation (SD) across four independent runs with different random seeds on the BraTS dataset. For the MVTec dataset, we follow standard practice and report the average performance over all defect subclasses as in prior work. Since most MVTec results are directly taken from the original papers, multiple-run evaluation under the same protocol is not always feasible.

**Results on BraTS Dataset**

The anomaly detection and localization results on the BraTS2021 dataset are summarized in Table 1. Our proposed method, DQN_AD demonstrates substantial improvements across all evaluated metrics, demonstrating the generalizability and effectiveness of the RL framework. The models are trained exclusively on the IXI dataset (healthy brain MRI volumes) and validated and tested on the BraTS2021 dataset (unhealthy brain MRI volumes with tumors). This cross-domain setup presents a more rigorous challenge than industrial benchmarks, due to the high heterogeneity of normal brain structures, domain shift, and the larger scale of the BraTS dataset, which often impairs the performance of existing methods.

Our proposed method, DQN_AD, achieves the highest AUROC scores at both the image level (88.7%) and the pixel level (96.7%), demonstrating strong anomaly detection and localization performance. Notably, our method also maintains high AUPRC (94.8%/66.2%) and [$DICE$] score (84.2%/49.7%), while achieving a balanced sensitivity (86.6%/63.0%) and specificity (85.2%/63.6%) at the optimal threshold. The qualitative results on BraTS 2021 dataset are shown in Figure 3, further illustrating the model's capacity to localize anomalies with high spatial precision. Compared to other methods, several approaches (e.g., BGAD, DRA) show higher sensitivity and DICE scores but substantially lower AUROC and specificity. This suggests that they may be over-sensitive to domain shifts and tend to over-segment by treating most regions as anomalies, resulting in higher false-positive rates. In contrast, DQN_AD maintains a favorable trade-off: while not achieving the absolute highest sensitivity, it preserves both specificity and localization accuracy, indicating that the model avoids excessive false positives while accurately detecting true anomalies. Moreover, DQN_AD shows consistent performance across both MRI modalities (T1 and T2) and across multiple metrics, with smaller standard deviations than many compared methods, reflecting its stability and robustness in generalization.

**Results on Industrial Dataset**

The results of anomaly detection and localization on the two industrial dataset are presented in Table 2. Our method demonstrates superior performance, achieving the highest average I-AUROC (99.8%, +0.4% over the semi-supervised SOTA) and P-AUROC (99.3%, +0.3%), as well as improvements of +0.2% I-AUROC and +0.8% P-AUROC over leading unsupervised competitors. Importantly, on notoriously difficult classes such as cable (100/99.2) and toothbrush (98.9, 98.6), our model still achieves outstanding results. On BTAD, our model maintains high performance (95.0/98.6), confirming our model's generalizability to different industrial anomaly domains. Qualitative examples on MVTec AD are shown in Figure 4.

**Ablation Study: Effect of the number of seen anomalies used**

As shown in Table 3, we explore the impact of the number of anomalies used. On industrial datasets, our model shows slightly lower performance compared to two semi-supervised methods BGAD and PRN with one seen anomaly. However, it outperforms them when using 10 seen anomalies. Our approach significantly outperforms other methods on BraTS 2021 with different number of seen anomalies from 1 – 300 with stable performance, demonstrating the robustness of our method across datasets with varying levels of imbalance.

**Ablation Study: Effect of subsampling size and frequency**

Subsampling is an important component in our framework as it acts like a scope to investigate the large dataset. The subsampling size and frequency determine the range of each observation and how quickly we move the scope around to investigate the dataset. Figure 5 shows that our algorithm is stable with fluctuations of approximately 0.7% in P_AUROC and 0.6% in I_AUROC when the size and frequency are varied within reasonable range. More figures and analysis are provided in Supplementary Material.

**Ablation Study: Backbone architecture, hierarchies and neighborhood size**

We evaluated our model using different backbones and tested its performance by extracting features from various hierarchy layers of WideResNet50 on the MVTec AD and BraTS 2021 dataset. The results, shown in Table 4 and Table S2, indicate that performance remains mostly stable across different layers and backbones. However, combining features from levels 2 and 3 of WideResNet50 yields the best performance.

As discussed in [1,10], the patch size $p$ controls the locality and global context of the locally aware patch-features extracted by the pre-trained extractor. We conducted the experiment to check the effect of patch size and the Figure S1 in the Supplementary Material shows consistent result as the PatchCore, $p = 3$ achieve the best result and balance between local and global context.

**Ablation Study: Comparison With classification**

We conducted additional experiments to highlight the limitations of simple classification, including challenges with large-scale datasets, class imbalance, overfitting, and distribution shifts. In contrast, our RL framework effectively addresses these issues, demonstrating its advantages. Further background, analysis, and experimental details are provided in the Supplementary Material.

**Ablation Study: Efficiency**

Our method ensures constant training time regardless of dataset size by efficiently sampling and exploring patches. On the larger BraTS 2021 dataset, training is even faster due to smaller image sizes. By focusing on critical points and skipping redundant ones like backgrounds, our approach accelerates convergence.

**Ablation Study: Optional Component Study**

As shown in Figure 6, on the MVTec AD dataset, each component contributes to improved performance, with RPAG showing the most significant boost, while PER and BS have smaller effects. However, on the BraTS 2021 dataset, PER, RPAG and BS have slightly negative impact on AUROC and $[DICE]$, likely due to increased false positives despite broader anomaly coverage as shown in the Figure S2 in Supplementary material. This may stem from overfitting caused by these techniques in the context of diverse normal distributions and testing distribution shifts, where RPAG generates overly subtle anomalies. These findings emphasize the importance of leveraging DQL techniques while applying feature learning methods judiciously, particularly in challenging medical datasets.

## Discussion

Anomaly detection in medical poses challenges due to blurred boundary between normal and abnormal, subtle lesion appearance, and domain shifts between training and testing data. In this work, we addressed these issues by formulating anomaly detection as a decision problem and training a DQN agent to learn a one-class decision boundary. Our method attains SOTA performance on both the BraTS 2021 brain MRI benchmark (T1 and T2 modality) and the MVTec AD/BTAD industrial surface-defect datasets, delivering balanced sensitivity and specificity essential for clinical utility, while maintaining high localization accuracy.

By comparing our reinforcement-learning framework against reconstruction-, synthesis-, and embedding-based unsupervised methods as well as recent semi-supervised approaches, we demonstrate our method can better accommodate class imbalance and overfitting problem. Unlike classification-based models that fail to handle these such extreme imbalanced cases and exist overfitting problem, even though our agent acts like a classification role, we shows that by selectively explores normal and anomalous feature pools, focusing training on challenging boundary cases and thereby improving efficiency and generalizability, our method is more like modeling the normal distribution while extending the boundary.

Theoretical analysis confirms that our algorithm meets convergence conditions for approximate Q-learning under static, fixed datasets. Our tailored exploration-exploitation scheme balances diversity (by sampling "farthest" normal points) with boundary refinement (by sampling "closest" misclassified point), mitigating overfitting to localized patterns.

Despite these advances, the current framework has limitations. First, maintaining a large replay buffer of patch features increases memory demands. Second, even though we are using locally-aware patch feature representation which can capture the local neighboring information, treating each patch independently still might overlooks spatial and contextual relationships that may be critical for detecting complex or collective anomalies. Finally, even though we are using DQN networks which work well on sequential data, but we treat the patches non-sequentially which remains space for improvement by using DRL algorithms.

Future work should explore memory-efficient replay strategies and incorporate inter- and intra-patch context or considering how to model data sequentially – potentially via graph-based models or spatiotemporal extensions – to capture structural and logical anomalies. Extending the framework to multimodal and dynamic imaging data, and conducting reader studies or prospective clinical trails, will be essential to fully realize the potential of reinforcement learning in medical anomaly detection.

In summary, our DQN-based semi-supervised anomaly detector provides a principled, generalizable and efficient solution that advances both medical and industrial applications. By combining reinforcement learning with robust feature learning, we offer a scalable approach for accurate, balanced anomaly detection and localization, laying the groundwork for more reliable AI-assisted diagnostic systems.

**References**

1. Liu Z, Zhou Y, Xu Y, Wang Z. Simplenet: A simple network for image anomaly detection and localization. In: *Proceedings of the IEEE/CVF Conference on Computer Vision and Pattern Recognition*. ; 2023:20402-20411.

2. Tien TD, Nguyen AT, Tran NH, et al. Revisiting reverse distillation for anomaly detection. In: *Proceedings of the IEEE/CVF Conference on Computer Vision and Pattern Recognition*. ; 2023:24511-24520.

3.  Ding C, Pang G, Shen C. Catching both gray and black swans: Open-set supervised anomaly detection. In: *Proceedings of the IEEE/CVF Conference on Computer Vision and Pattern Recognition*. ; 2022:7388-7398.

4.  Zhang H, Wu Z, Wang Z, Chen Z, Jiang YG. Prototypical residual networks for anomaly detection and localization. In: *Proceedings of the IEEE/CVF Conference on Computer Vision and Pattern Recognition*. ; 2023:16281-16291.

5.  Fährmann D, Jorek N, Damer N, Kirchbuchner F, Kuijper A. Double deep q-learning with prioritized experience replay for anomaly detection in smart environments. *IEEE Access*. 2022;10:60836-60848.

6.  Pang G, van den Hengel A, Shen C, Cao L. Toward deep supervised anomaly detection: Reinforcement learning from partially labeled anomaly data. In: *Proceedings of the 27th ACM SIGKDD Conference on Knowledge Discovery & Data Mining*. ; 2021:1298-1308.

7.  Isensee F, Schell M, Pflueger I, et al. Automated brain extraction of multisequence MRI using artificial neural networks. *Human brain mapping*. 2019;40(17):4952-4964.

8.  Jenkinson M, Smith S. A global optimisation method for robust affine registration of brain images. *Medical image analysis*. 2001;5(2):143-156.

9.  Jenkinson M, Bannister P, Brady M, Smith S. Improved optimization for the robust and accurate linear registration and motion correction of brain images. *Neuroimage*. 2002;17(2):825-841.

10. Roth K, Pemula L, Zepeda J, Schölkopf B, Brox T, Gehler P. Towards total recall in industrial anomaly detection. In: *Proceedings of the IEEE/CVF Conference on Computer Vision and Pattern Recognition*. ; 2022:14318-14328.

11. Mnih V, Kavukcuoglu K, Silver D, et al. Playing atari with deep reinforcement learning. *arXiv preprint arXiv:13125602*. Published online 2013.

12. Mnih V, Kavukcuoglu K, Silver D, et al. Human-level control through deep reinforcement learning. *nature*. 2015;518(7540):529-533.

13. Guo J, Lu S, Jia L, Zhang W, Li H. Encoder-Decoder Contrast for Unsupervised Anomaly Detection in Medical Images. *IEEE Transactions on Medical Imaging*. Published online 2023:1-1. doi:10.1109/TMI.2023.3327720

14. Kascenas A, Sanchez P, Schrempf P, et al. The role of noise in denoising models for anomaly detection in medical images. *Medical Image Analysis*. 2023;90:102963. doi:10.1016/j.media.2023.102963

15. Zhao Y, Ding Q, Zhang X. AE-FLOW: autoencoders with normalizing flows for medical images anomaly detection. In: *The Eleventh International Conference on Learning Representations*. ; 2022.

16. Zhang Z, Mohsenzadeh Y. Efficient Slice Anomaly Detection Network for 3D Brain MRI Volume. *arXiv preprint arXiv:240815958*. Published online 2024.

17. Yao X, Li R, Zhang J, Sun J, Zhang C. Explicit boundary guided semi-push-pull contrastive learning for supervised anomaly detection. In: *Proceedings of the IEEE/CVF Conference on Computer Vision and Pattern Recognition*. ; 2023:24490-24499.

18. Pang G, Ding C, Shen C, Hengel A van den. Explainable deep few-shot anomaly detection with deviation networks. *arXiv preprint arXiv:210800462*. Published online 2021.

**Tables**

**Table 1: Quantitative performance on BraTS 2021 T1 and T2 modality with 10 available anomalies.**

|  | AUROC | AUPRC | Sensitivity | Specificity | [*DICE*] |
|---|---|---|---|---|---|
| **T1 modality** | | | | | |
| | | | Unsupervised | | |
| SimpleNet | 70.3±1.3/88.7±1.1 | 80.9±1.0/26.2±1.1 | 95.9±0.9/47.9±0.4 | 70.5±0.8/26.3±0.8 | 39.5±3.5/**53.8±1.3** |
| EDC | 74.7±2.1/89.5±2.2 | 83.4±2.2/27.6±4.8 | 94.8±0.2/53.1±2.4 | 72.2±0.8/29.1±4.7 | 81.9±0.5/34.0±4.5 |
| AE-FLOW | 48.4±2.1/70.4±1.7 | 69.5±0.2/4.6±0.3 | 99.9±0.1/**77.7±0.6** | 68.1±0.1/10.0±0.6 | 81.0±0.0/5.5±0.3 |
| DAE | 57.8±0.4/60.6±0.0 | 74.7±0.8/10.8±0.0 | **100±0.0**/76.0±0.4 | 67.5±0.0/8.4±0.0 | 80.6±0.0/11.8±0.1 |
| PatchCore | 76.9±0.5/93.9±0.2 | 85.0±0.3/46.0±0.9 | 94.1±0.3/56.2±0.2 | 73.4±0.2/45.2±1.0 | 82.4±0.1/47.7±0.7 |
| RD++ | 77.7±1.2/94.2±0.3 | 85.6±1.0/43.3±2.1 | 92.8±0.9/60.2±0.7 | 74.3±1.0/42.5±1.9 | 82.5±0.3/47.2±1.5 |
| SSNet | **78.5±0.5/94.4±0.1** | **86.4±0.4/49.1±0.7** | 92.6±0.7/62.4±0.1 | **75.2±0.6/47.0±0.7** | 83.0±0.1/50.1±0.5 |
| | | | Semi-Supervised | | |
| BGAD | 53.2±2.8/80.1±0.9 | 70.1±1.8/9.7±0.5 | **100±0.0**/59.9±1.7 | 67.4±0.0/11.0±0.5 | 80.5±0.0/16.8±0.6 |
| DRA | 69.5±3.7/- | 81.6±2.1/- | 97.5±2.5/- | 69.0±1.6/- | 80.5±0.0/- |
| Ours | **84.0±1.4**/93.8±0.6 | **92.5±0.7/51.3±1.1** | 88.5±0.7/53.8±0.1 | **78.3±2.1/50.7±1.9** | 75.8±1.7/**30.6±0.6** |
| **T2 modality** | | | | | |
| | | | Unsupervised | | |
| SimpleNet | 75.3±0.8/90.6±1.2 | 85.1±0.3/32.2±2.1 | 93.8±0.8/48.9±1.7 | 71.9±1.1/31.9±1.6 | 36.5±1.5/52.1±1.0 |
| EDC | 82.6±1.3/90.0±3.8 | 89.6±1.2/48.2±6.6 | 89.2±1.1/58.7±2.8 | 79.0±1.2/36.2±7.8 | **83.8±0.2**/40.1±6.5 |
| AE-FLOW | 56.6±0.2/82.5±1.1 | 70.9±0.1/13.0±0.7 | 98.9±0.4/52.2±0.8 | 68.8±0.5/14.8±0.8 | 81.1±0.2/20.3±0.9 |
| DAE | 72.0±0.9/64.7±0.0 | 84.5±0.4/35.0±0.5 | **100±0.0**/50.3±0.1 | 67.6±0.0/44.6±0.1 | 80.7±0.0/32.3±0.2 |
| PatchCore | 82.6±0.4/96.1±0.1 | 90.2±0.3/**58.0±0.6** | 90.2±0.7/62.2±0.3 | 77.5±0.9/**55.8±0.6** | 83.4±0.2/**57.6±0.5** |
| RD++ | 82.4±0.6/**96.2±0.2** | 89.9±0.4/52.3±1.7 | 90.2±0.5/63.0±0.4 | 78.1±0.7/49.7±1.4 | 83.7±0.3/53.8±1.0 |
| SSNet | **83.4±0.3/96.2±0.1** | 90.5±0.1/57.7±0.4 | 88.4±0.6/**65.9±0.1** | **79.5±0.7**/53.7±0.3 | 83.7±0.2/56.4±0.3 |
| | | | Semi-Supervised | | |
| BGAD | 55.0±0.8/85.0±0.7 | 70.3±0.6/16.4±1.2 | **99.9±0.1**/54.7±2.9 | 67.5±0.1/17.8±1.5 | 80.5±0.0/23.5±1.5 |
| DRA | 61.4±3.1/- | 77.4±2.3/- | 100±0.0/- | 67.4±0.0/- | 80.5±0.0/- |
| Ours | **88.7**±1.0/**96.7**±0.4 | **94.8**±0.5/**66.2**±3.2 | 86.6±0.9/**63.0**±2.6 | 85.2±1.8/**63.6**±2.9 | 84.2±1.7/49.7±4.7 |

Results are reported with 10 available anomalies. Metrics are shown as image-level anomaly detection / pixel-level localization (mean ± SD). SSNet refers to the 3D method SimpleSliceNet method. Ours refers to our pure RL base framework DQN_Base without any optional components. Metrics include AUROC (area under ROC curve), AUPRC (area under precision-recall curve), Sensitivity, Specificity, and maximum DICE score [*DICE*] are computed at the optimal threshold that yield the highest DICE score. Results are reported as mean ± SD over 4 runs. Best values are bolded.

**Table 2: Quantitative performance on industrial datasets with 10 available anomalies.**

| | Unsupervised | | | | Semi-Supervised | | | |
|---|---|---|---|---|---|---|---|---|
| | SimpleNet | PatchCore | RD++ | BGAD | DevNet | DRA | PRN | Ours |
| carpet | 99.7/98.2 | 99.1/99.0 | **100**/99.2 | **100**/99.5 | 82.5/97.2 | 92.5/98.2 | 99.7/99.0 | **100**/**99.7** |
| grid | 99.7/98.8 | 97.3/98.7 | **100**/99.3 | **100**/99.3 | 90.6/87.9 | 98.6/86.0 | 99.4/98.4 | **100**/**99.4** |
| leather | **100**/99.2 | **100**/99.3 | **100**/99.4 | **100**/99.7 | 92.2/94.2 | 98.9/93.8 | **100**/99.7 | **100**/**99.9** |
| tile | 99.8/97.0 | 99.3/95.8 | 99.7/96.6 | **100**/99.0 | 99.9/92.7 | **100**/92.3 | **100**/99.6 | **100**/99.1 |
| wood | **100**/94.5 | 99.6/95.1 | 99.3/95.8 | 98.8/97.3 | 97.9/86.4 | 99.1/82.9 | **100**/97.8 | 99.9/**98.2** |
| bottle | **100**/98.0 | **100**/98.6 | **100**/98.8 | **100**/98.9 | 99.7/93.9 | **100**/91.3 | **100**/99.4 | **100**/99.1 |
| cable | 99.9/97.6 | 99.9/98.5 | 99.2/98.4 | 89.2/96.7 | 71.9/91.8 | 94.2/86.6 | 98.9/98.8 | **100**/**99.2** |
| capsule | 97.7/98.9 | 98.0/99.0 | 99.0/98.8 | 82.7/98.1 | 99.7/91.1 | 95.1/89.3 | 98.0/98.5 | **99.9**/**99.2** |
| hazelnut | **100**/97.9 | **100**/98.7 | **100**/99.2 | **100**/99.5 | 98.8/77.8 | **100**/89.6 | **100**/**99.7** | **100**/99.6 |
| metal nut | **100**/98.8 | 99.9/98.3 | **100**/98.1 | 99.6/99.4 | 87.1/82.6 | 99.1/79.5 | **100**/**99.7** | **100**/99.2 |
| pill | 99.0/98.6 | 97.5/97.6 | 98.4/98.3 | 98.2/99.3 | 97.2/60.3 | 88.3/84.5 | **99.3**/99.5 | **99.3**/99.4 |
| screw | 98.2/99.3 | 98.2/99.5 | 98.9/**99.7** | 87.2/98.6 | 79.2/84.6 | **99.5**/54.0 | 95.9/97.5 | 98.5/**99.7** |
| toothbrush | 99.7/98.5 | **100**/98.6 | **100**/99.1 | 87.5/98.1 | 89.1/56.0 | 87.5/75.5 | **100**/**99.6** | 98.9/98.6 |
| transistor | **100**/97.6 | 99.9/96.5 | 98.5/94.3 | 95.3/94.9 | 99.1/56.0 | 88.3/79.1 | 99.7/98.4 | **100**/**99.1** |
| zipper | 99.9/98.9 | 99.5/98.9 | 98.6/98.8 | 99.5/99.0 | 99.1/93.7 | 99.7/96.9 | 99.7/98.8 | **100**/**99.5** |
| MVTec_AD | 99.6/98.1 | 99.2/98.1 | 99.4/98.3 | 95.9/98.5 | 92.2/85.3 | 96.1/85.3 | 99.4/99.0 | **99.8**/**99.3** |
| BTAD | **95.7**/97.2 | 92.6/96.9 | 95.6/97.4 | 94.6/**98.6** | -/- | 94.2/75.4 | 94.7/97.1 | 95.0/**98.6** |

Rows 1-15 report image-level / pixel-level AUROC (%) for each of the 15 defect subclasses in MVTec AD. The 16th and 17th rows give the mean AUROC across their subclasses. Unsupervised methods (SimpleNet, PatchCore, RD++, BGAD) and semi-supervised methods (DevNet, DRA, PRN, Ours) are grouped by columns. All approaches' results are cited from their original publications. Best values are bolded.

**Table 3: Performance on three datasets with different number of seen anomalies used.**

| Dataset | $|D^a|$ | BGAD | DRA | PRN | Ours |
|---|---|---|---|---|---|
| BraTS2021 | 1 | 53.6±1.9/80.4±1.3 | 56.8±3.6/- | -/- | **78.6±2.2/92.5±1.1** |
| | 10 | 55.0±0.8/85.0±0.7 | 61.4±3.1/- | -/- | **88.7±1.0/96.7±0.4** |
| | 50 | 55.8±2.2/87.5±1.0 | 69.3±2.9/- | -/- | **89.1±0.6/97.5±0.1** |
| | 100 | 56.0±6.3/91.5±1.4 | 66.6±4.9/- | -/- | **90.0±1.2/97.7±0.2** |
| | 200 | 53.8±3.3/82.2±4.8 | 68.9±4.4/- | -/- | **89.7±1.1/97.4±0.3** |
| | 300 | 57.6±5.8/84.9±2.9 | 68.4±4.7/- | -/- | **89.9±0.7/97.6±0.3** |
| MVTec AD | 1 | 89.8/95.2 | 88.9/78.8 | **98.8/98.3** | 98.3/96.7 |
| | 10 | 95.9/98.5 | 96.1/85.3 | 99.4/99.0 | **99.8/99.3** |
| BTAD | 1 | **94.2/98.2** | 83.6/- | -/- | 92.9/97.8 |
| | 10 | 94.6/98.6 | 94.2/- | -/- | **95.0/98.6** |

Results (I-AUROC/P-AUROC, mean±standard deviation) are shown for semi-supervised methods – BGAD, DRA, PRN, and Ours – on three datasets as the size of the labeled anomaly pool varies. For BraTS2021, $|D^a|$ ranges from 1 to 300; for MVTec AD and BTAD, results are reported at $|D^a|$ =1 and 10. PRN code is not available ("-"). Results are reported as mean ± SD over 4 runs. Results on BraTS2021 are reported as mean ± SD over 4 runs. Best values are bolded.

**Table 4: Performance on MVTec AD and BraTS2021 under different hierarchy levels.**

| Level 1 | Level 2 | Level 3 | MVTec_AD | BraTS 2021 (AUROC) | BraTS 2021 (AUPRC) |
|---|---|---|---|---|---|
| Y | | | 97.3/93.3 | 82.3±2.5/95.1±0.5 | 91.4±1.0/57.8±5.6 |
| | Y | | 97.8/96.1 | 87.9±0.5/97.0±0.3 | 94.1±0.3/64.9±1.7 |
| | | Y | 96.1/95.0 | 82.5±1.7/91.9±2.0 | 91.2±1.1/47.5±5.3 |
| Y | Y | | 98.2/96.2 | 88.1±0.6/96.9±0.8 | 94.2±0.3/65.6±4.5 |
| Y | | Y | 98.1/**97.0** | 86.7±0.4/**97.3±0.2** | 93.7±0.3/**67.8±2.5** |
| | Y | Y | **98.3**/96.7 | **88.7±1.0**/96.7±0.4 | **94.8±0.5**/66.2±3.2 |
| Y | Y | Y | 97.8/96.9 | 87.1±0.7/97.1±0.5 | 93.9±0.4/66.8±4.3 |

Results are shown for models using different combinations of hierarchy levels. "Y" indicates that the corresponding level is included. Performance is evaluated on the MVTec AD dataset (I-AUROC/P-AUROC) and on BraTS 2021 (image-/pixel-level AUROC and AUPRC). Each row corresponds to one combination of hierarchy levels. Results on BraTS2021 are reported as mean ± SD over 4 runs. Best values are bolded.

# Figures

**Figure 1. Pseudo Algorithm of Anomaly Detection using a deep Q-network.**

Pseudocode of the reinforcement learning-based anomaly detection algorithm. The agent interacts with a mixed feature set sampled from labeled and unlabeled data. A DQN is trained to optimize anomaly identification guided by reward feedback. $S^a$ and $S^u$ denote normal and abnormal features respectively.

**Figure 2. Overview of the proposed DQN-based anomaly detection framework.**

The pipeline consists of three modules – Environment (left), Agent (top right), and Replay Buffer (bottom right) – each enclosed by dashed borders. In the Environment, the current state $s_t$ is passed to the Agent, which returns an action $a_t$; the Environment then samples the next state $s_{t+1}$ from normal ($S^u$) or anomalous ($S^a$) feature pools and computing the reward $r_t$. Dark blue arrows (Interaction) trace this online loop of state sampling, action selection, reward computation, and transition storage. Teal arrows (Training) indicate the offline learning loop: sampling transitions ($s_t, a_t, r_t, s_{t+1}$) from the Replay Buffer to update the Q-network parameters and periodically synchronize the target Q-network.

**Figure 3. Qualitative comparison of anomaly localization on the BraTS 2021 dataset.**

The representative MRI slices are split into two horizontal blocks. Rows corresponding to different slices and columns (left to right) show: (column 1) the original T2-weighted image, (column 2) the ground-truth anomaly mask, (column 3-7) localization outputs of baseline methods, and (column 8) the localization output of our model. All binary masks were produced by thresholding each model's anomaly score map at its best F1-score threshold.

**Figure 4. Qualitative anomaly localization results on the MVTec AD dataset.**

Fifteen defect classes from MVTec AD are shown, one example per class, arranged in three horizontal blocks for compact layout. Within each block, rows correspond to different classes and columns (left to right) display: (column 1) the original surface image, (column 2) the ground-truth defect mask, (column 3-4) localization masks from baseline methods, and (column 4) the localization mask produced by our model variant DQN_All. All binary masks were obtained by thresholding each model's anomaly score map at its optimal F1-score threshold.

**Figure 5. Effect of subsampling frequency and pool size on model performance**

(a) Average pixel-level AUROC as a function of the sampling frequency for three subsamples sizes $(1 \times 10^4, 5 \times 10^4, 1 \times 10^5)$. (b) Average image-level AUROC under the same conditions. Frequency refers to the interval between subsampling steps (e.g., a frequency of 100 means subsampling every 100 steps).

**Figure 6. Component-wise ablation on MVTec AD and BraTS 2021 anomaly detection performance.**

(a) Image- and pixel-level AUROC on BraTS 2021 with ten labeled anomalies (mean $\pm$ SD). (b) Image- and pixel-level maximum Dice score on BraTS 2021 (mean $\pm$ SD). (c) Image- and pixel-level AUROC on MVTec AD with one labeled anomaly. Bars compare eight variants of our DQN-based framework: the baseline (DQN_Base); DQN-Base augmented with prioritized experience replay (+PER), data augmentation (+RPAG), or boundary supervision (+BS); the full model with all components (DQN_All); and ablations removing each component from the full model (-PER, -RPAG, -BS). Results on BraTS2021 are reported as over 4 runs.

Algorithm 1. Anomaly Detection using DQN

**Input:** $D = \{D^a, D^u\}$ – training data

**Parameter:** $\theta$ – action-value network parameters

**Output:** $Q(s, a, \theta^*)$ – trained action-value network

1. Initialize action-value network $Q$ with random weights $\theta$
2. Initialize target action-value function $Q$
3. Initialize experience replay buffer $B$ with size $M$
4. Initialize subset of features $\{S^a, S^u\}$ from $D$
5. **for** i = 1 to *n_episodes* **do**:
6.    Obtain initial state $s_1 \sim U(S^u)$
7.    **for** t = 1 to *n_steps* **do**:
8.       With probability $\epsilon$, select random action $a_t \in \{a^0, a^1\}$
9.       Otherwise select $a_t = \arg\max_a Q(s_t, a; \theta)$
10.      With probability $p$, sample $s_{t+1} \sim S^a$; else $s_{t+1} \sim S^u$
11.      Compute reward $r_t$
12.      Store experience $(s_t, a_t, r_t, s_{t+1})$ in $B$
13.      Sample minibatch of experience records $(s', a', r', s'_{t+1})$ from $B$
14.      Compute $Loss = r' + \gamma \max'_a Q(s'_{t+1}, a') - Q(s', a')$
15.      Update $\theta$ by minimizing Loss
16.      Update $\hat{Q} \leftarrow Q$ every $K$ steps
17.      Update $\{S^a, S^u\}$ every $T$ steps
18.    **end for**
19. **end for**
20. **Return trained $Q$**

**Figure 1. Pseudo Algorithm of Anomaly Detection using a deep Q-network.**

Pseudocode of the reinforcement learning-based anomaly detection algorithm. The agent interacts with a mixed feature set sampled from labeled and unlabeled data. A DQN is trained to optimize anomaly identification guided by reward feedback. $S^a$ and $S^u$ denote normal and abnormal features respectively.

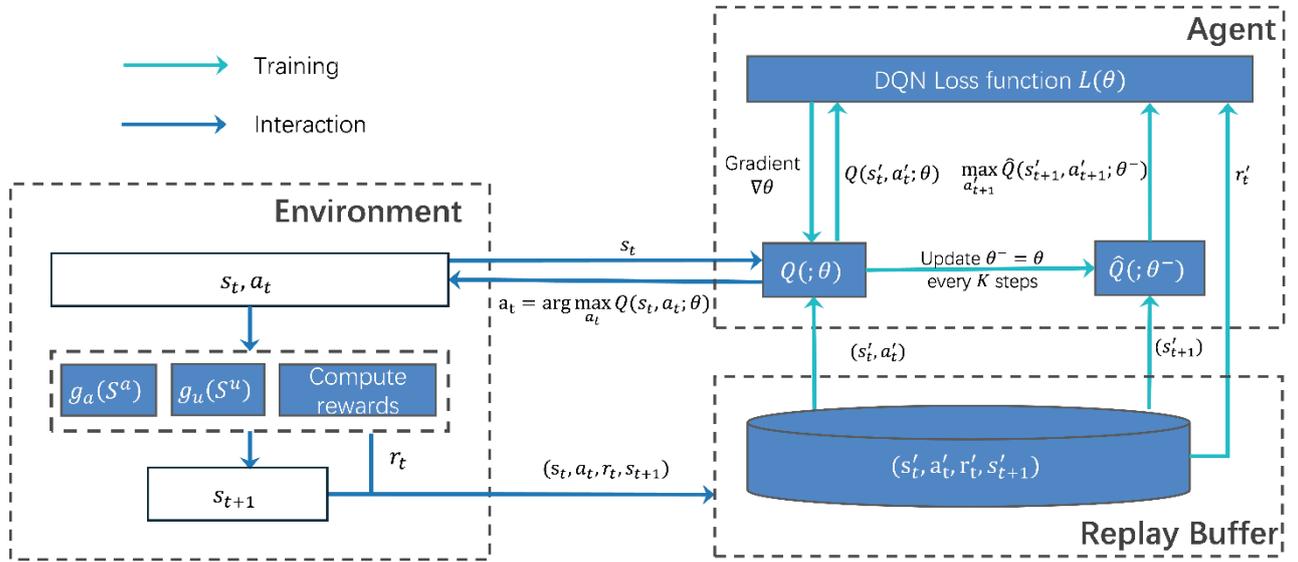

**Figure 2. Overview of the proposed DQN-based anomaly detection framework.**

The pipeline consists of three modules – Environment (left), Agent (top right), and Replay Buffer (bottom right) – each enclosed by dashed borders. In the Environment, the current state $s_t$ is passed to the Agent, which returns an action $a_t$; the Environment then samples the next state $s_{t+1}$ from normal ($S^u$) or anomalous ($S^a$) feature pools and computing the reward $r_t$. Dark blue arrows (Interaction) trace this online loop of state sampling, action selection, reward computation, and transition storage. Teal arrows (Training) indicate the offline learning loop: sampling transitions ($s_t, a_t, r_t, s_{t+1}$) from the Replay Buffer to update the Q-network parameters and periodically synchronize the target Q-network.

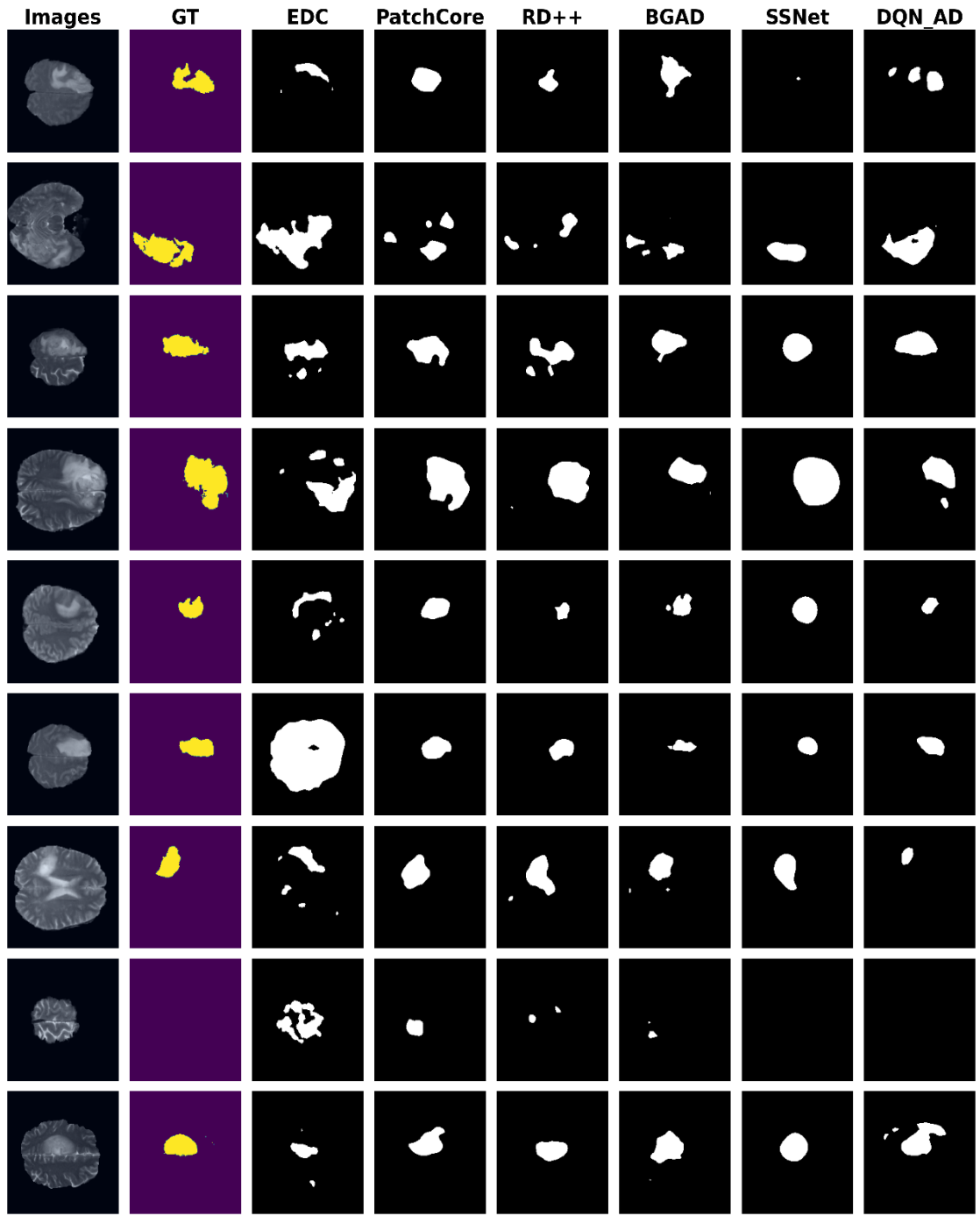

**Figure 3. Qualitative comparison of anomaly localization on the BraTS 2021 dataset.**

The representative MRI slices are split into two horizontal blocks. Rows corresponding to different slices and columns (left to right) show: (column 1) the original T2-weighted image, (column 2) the ground-truth anomaly mask, (column 3-7) localization outputs of baseline methods, and (column 8) the localization output of our model. All binary masks were produced by thresholding each model's anomaly score map at its best F1-score threshold.

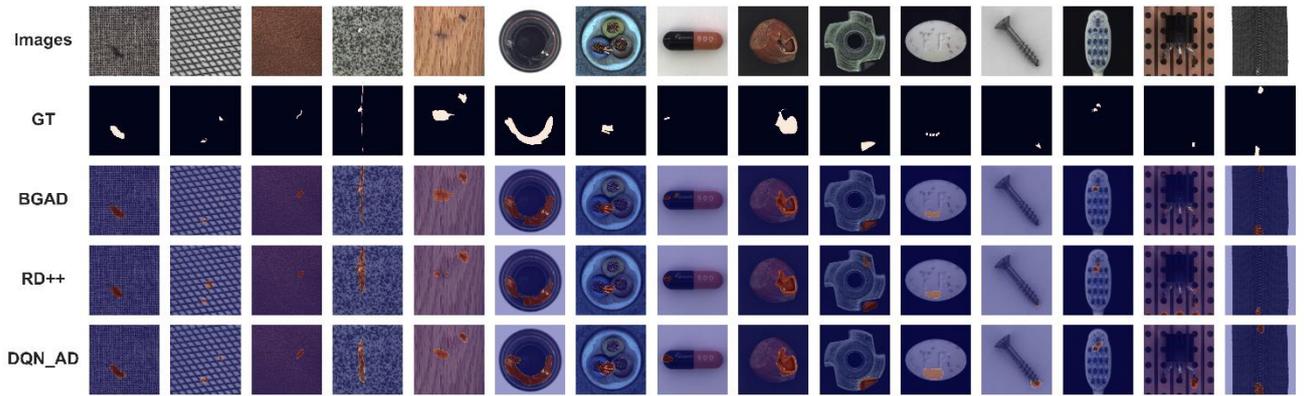

**Figure 4. Qualitative anomaly localization results on the MVTec AD dataset.**

Fifteen defect classes from MVTec AD are shown, one example per class, arranged in three horizontal blocks for compact layout. Within each block, rows correspond to different classes and columns (left to right) display: (column 1) the original surface image, (column 2) the ground-truth defect mask, (column 3-4) localization masks from baseline methods, and (column 4) the localization mask produced by our model variant DQN_All. All binary masks were obtained by thresholding each model's anomaly score map at its optimal F1-score threshold.

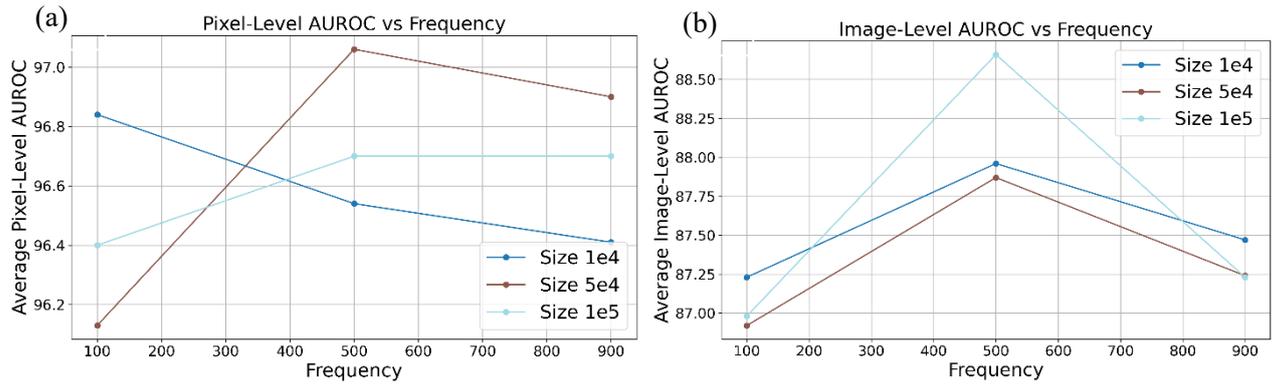

**Figure 5. Effect of subsampling frequency and pool size on model performance**

(a) Average pixel-level AUROC as a function of the sampling frequency for three subsamples sizes $(1 \times 10^4, 5 \times 10^4, 1 \times 10^5)$. (b) Average image-level AUROC under the same conditions. Frequency refers to the interval between subsampling steps (e.g., a frequency of 100 means subsampling every 100 steps).

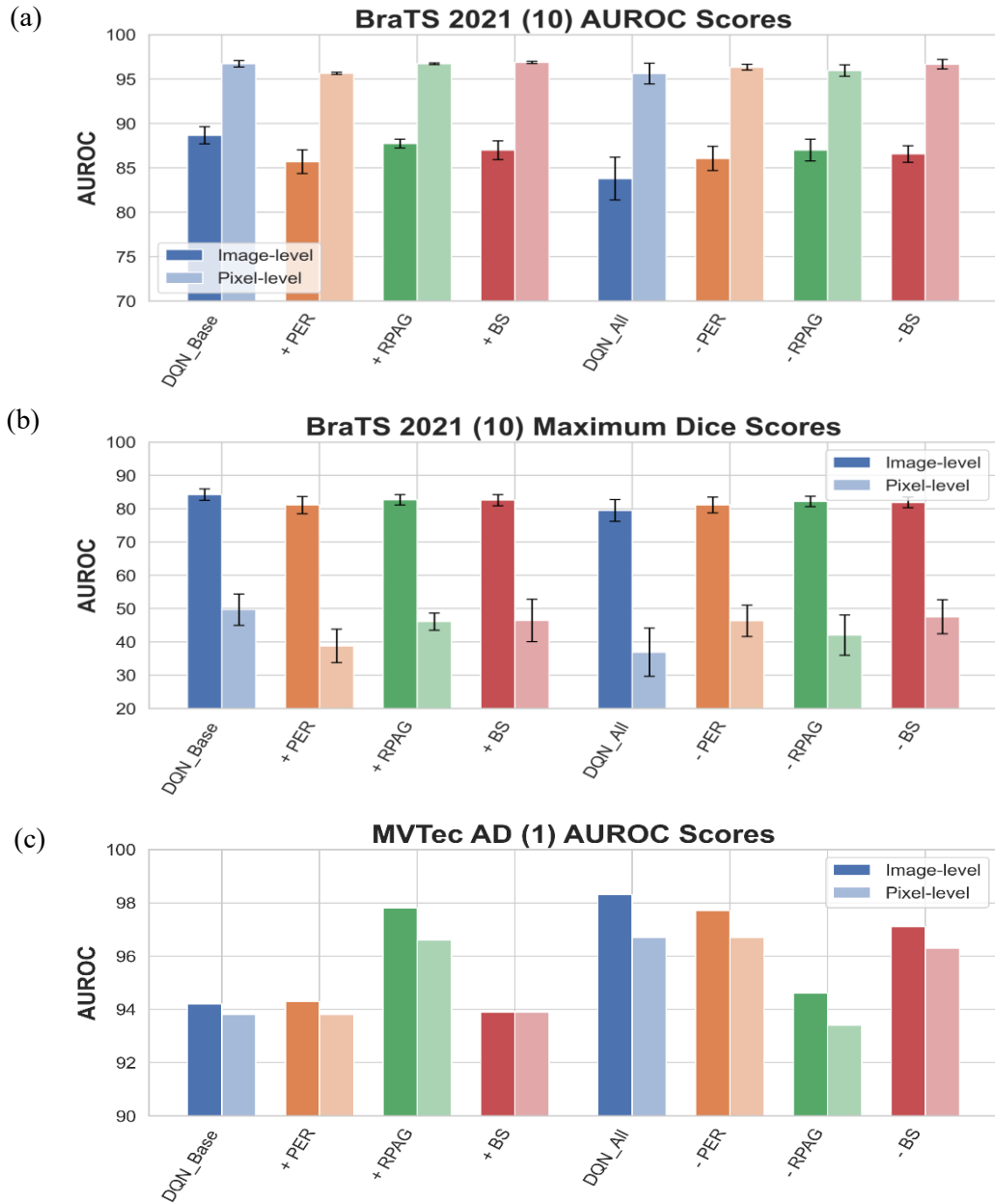

**Figure 6. Component-wise ablation on MVTec AD and BraTS 2021 anomaly detection performance.**

(a) Image- and pixel-level AUROC on BraTS 2021 with ten labeled anomalies (mean ± SD). (b) Image- and pixel-level maximum Dice score on BraTS 2021 (mean ± SD). (c) Image- and pixel-level AUROC on MVTec AD with one labeled anomaly. Bars compare eight variants of our DQN-based framework: the baseline (DQN_Base); DQN-Base augmented with prioritized experience replay (+PER), data augmentation (+RPAG), or boundary supervision (+BS); the full model with all components (DQN_All); and ablations removing each component from the full model (-PER, -RPAG, -BS). Results on BraTS2021 are reported as over 4 runs.

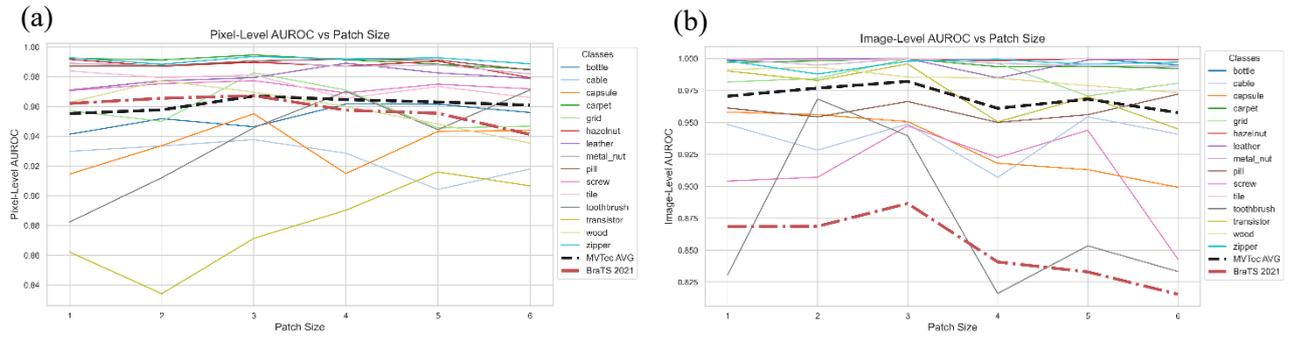

**Figure S1. Effect of patch size of pretrained extractor on anomaly detection performance.**

(a) Pixel-level AUROC as a function of patch size for each of the 15 MVTec AD classes (colored solid lines), the average over all MVTec classes (black dashed line), and BraTS 2021 results (red dash–dot line). (b) Image-level AUROC under the same conditions. In both panels, the x-axis denotes the side length of the square patch, and the y-axis shows the corresponding AUROC; BraTS 2021 values are plotted for reference against industrial benchmarks.

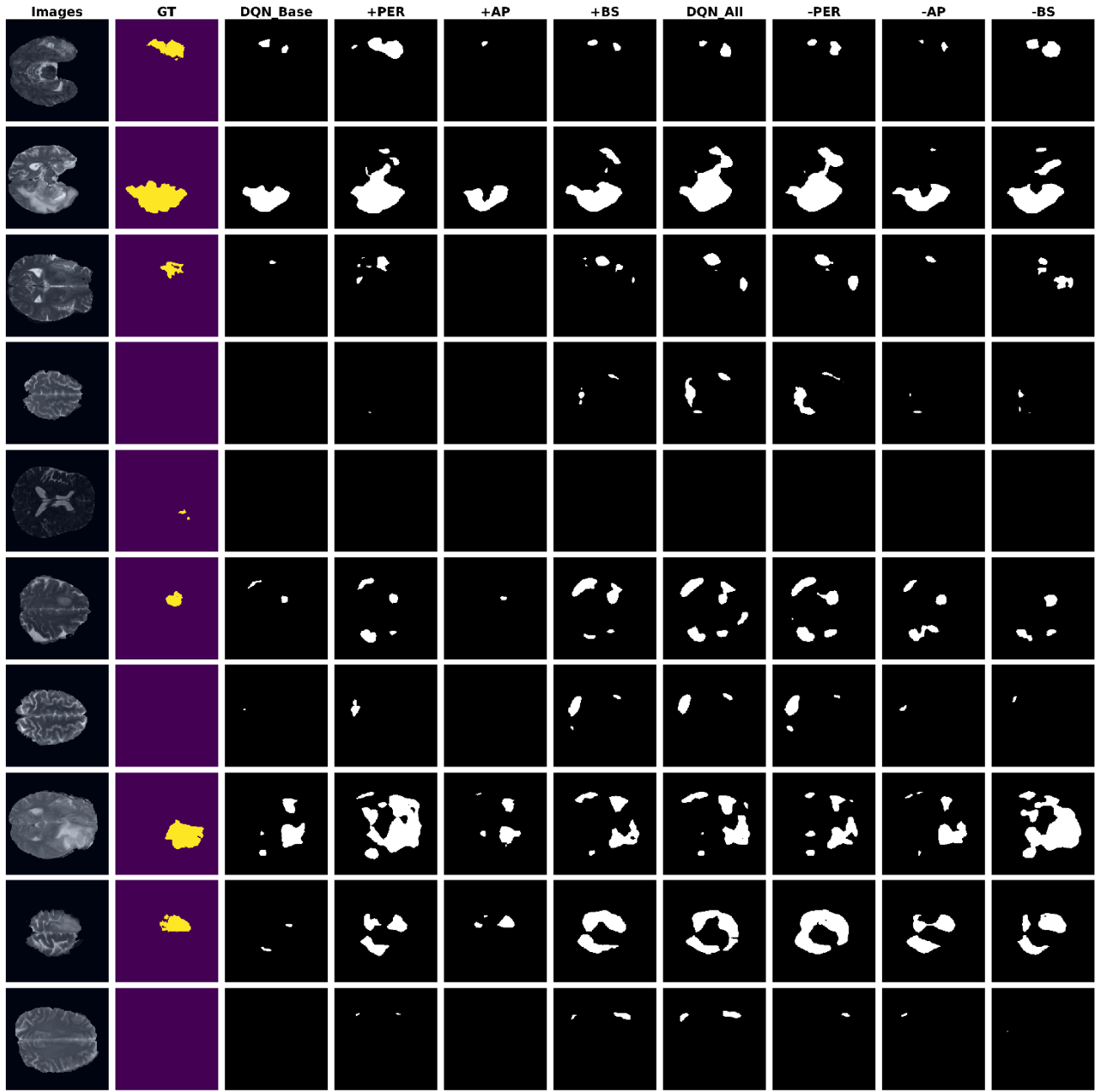

**Figure S2. Qualitative localization results for the component-wise ablation study.**

Representative MRI slices (first column) and their ground-truth anomaly masks (second column) are shown alongside binary masks produced by eight model variants (columns 3-10). All binary masks were obtained by thresholding each variant's anomaly score map at its optimal F1-score threshold.

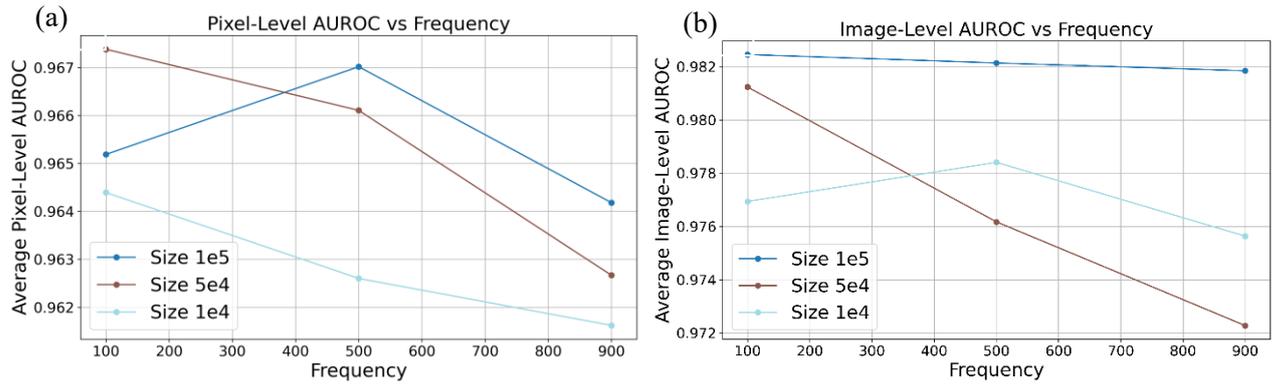

**Figure S3. Effect of subsampling frequency and pool size on model performance on MVTec AD**

(a) Average pixel-level AUROC as a function of the sampling frequency for three subsamples sizes $(1 \times 10^4, 5 \times 10^4, 1 \times 10^5)$. (b) Average image-level AUROC under the same conditions. Frequency refers to the interval between subsampling steps (e.g., a frequency of 100 means subsampling every 100 steps).

**Supplementary Materials**

**Implementation Details**

Images from the MVTec AD and BTAD datasets were resized to $256 \times 256$, while volumes from IXI and BraTS 2021 were preprocessed and cropped to $192 \times 192 \times 96$.

In our experiments, all backbone architectures were pre-trained on ImageNet. For other models, we followed the original implementations specified in their respective papers. Our method utilized layers 2 and 3 of WideResnet50 as the backbone. Feature vectors were extracted with dimensions of 1024 for industrial datasets and 512 for medical datasets. As detailed in **Ablation Study: Optional Component Study** section, we used the DQN framework with three optional components for industrial datasets, referred to as DQN_All. For medical datasets, we implemented the DQN framework without components, denoted as DQN_Base. The action-value network was a simple multi-layer neural network with hidden size of [512, 256, 128] for industrial datasets and [256, 128] for medical datasets. The probability of actions output by the agent was normalized, with the probability of being anomalous, $\frac{Q^*(s,a^1)}{\sum_{a'} Q^*(s,a')}$, used as anomaly score.

Each intermediate layer in the action network is followed by a Rectified Linear Unit activation layer. Finally, a softmax activation layer is applied as the final activation layer. The DQN undergoes training for 40,000 steps, with an initial warm-up phase of 2,000 steps. The target network is updated every $K = 5,000$ steps. Each episode consists of $n_{steps} = 2,000$ steps, and an episode terminates only after all 2,000 steps are executed. The parameter $\epsilon$ is exclusively employed during training, gradually decreasing from 1.0 to 0.1 over 1,000 steps. The other parameters are listed in the Table S1. These hyperparameters are adjusted through experimentation, and the final values are determined based on quantitative and qualitative results on both industrial and medical datasets.

| Table S1. Key hyperparameters of the DQN-based anomaly detection framework. | | |
|---|---|---|
| Parameter | Value | Info |
| M | 10,000 | Size of the experience replay buffer |
| T | 500 | Frequency of updating normal/abnormal feature subsets (every T steps) |
| N | 80 | Number of images used for normal feature subsampling in each subsampling step |
| $|S^u|$ | 100,000 | The subsampling size, the number of normal features saved per step |

The table summarizes the main parameters used in our experiments.

**Ablation Study: Backbone Study**

| Table S2. Effect of backbone architecture on anomaly detection performance | | | |
|---|---|---|---|
| | ResNet18 | ResNet50 | WideResNet50 |
| MVTec AD | 97.8/96.3 | 98.3/96.0 | **98.3/96.7** |
| BraTS 2021 (AUROC) | 87.0±1.6/96.5±0.6 | 86.4±0.7/96.2±0.4 | **88.7±1.0/96.7±0.4** |
| BraTS 2021 (AUPRC) | 93.8±1.0/62.8±3.5 | 93.7±0.4/63.8±2.1 | **94.8±0.5/66.2±3.2** |

Results are reported as I-AUROC/P-AUROC on the MVTec AD dataset, and as image-level / pixel-level AUROC and AURPC (mean ± SD) on the BraTS 2021 dataset. Three backbone networks – ResNet18, ResNet50, and WideResNet50 – are compared.

**Ablation Study: Patch Size Study**

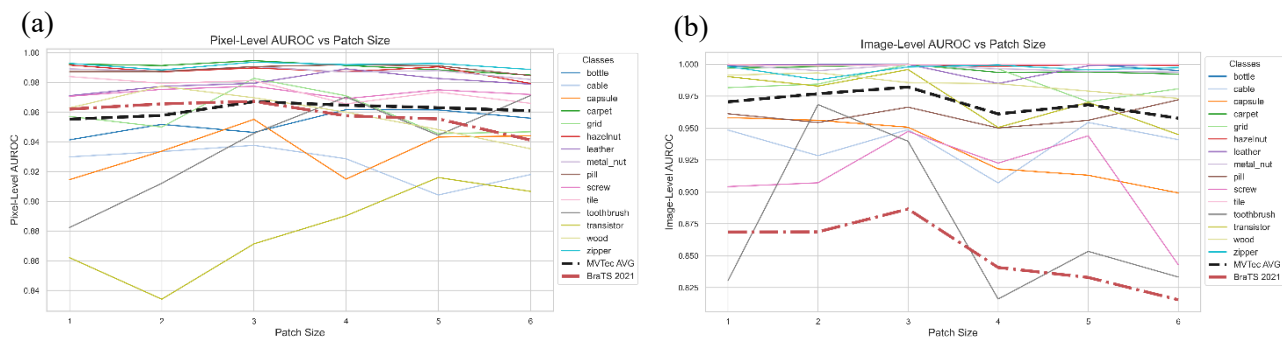

**Figure S1. Effect of patch size of pretrained extractor on anomaly detection performance.**

(a) Pixel-level AUROC as a function of patch size for each of the 15 MVTec AD classes (colored solid lines), the average over all MVTec classes (black dashed line), and BraTS 2021 results (red dash–dot line). (b) Image-level AUROC under the same conditions. In both panels, the x-axis denotes the side length of the square patch, and the y-axis shows the corresponding AUROC; BraTS 2021 values are plotted for reference against industrial benchmarks.

**Ablation Study: Qualitative Results of Component Study**

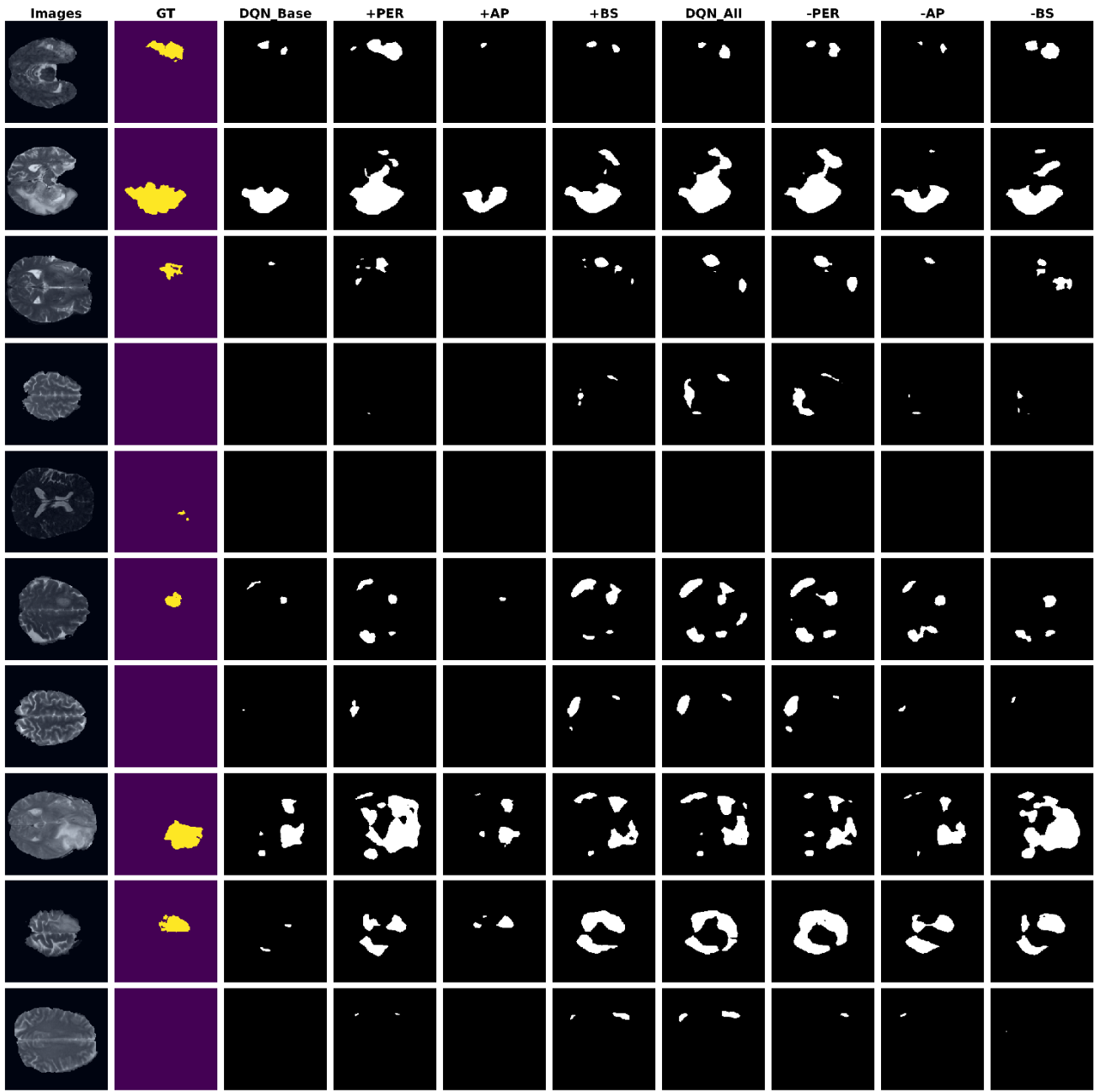

**Figure S2. Qualitative localization results for the component-wise ablation study.**

Representative MRI slices (first column) and their ground-truth anomaly masks (second column) are shown alongside binary masks produced by eight model variants (columns 3-10). All binary masks were obtained by thresholding each variant's anomaly score map at its optimal F1-score threshold.

**Ablation Study: Study on Subsampling size and frequency on MVTec AD**

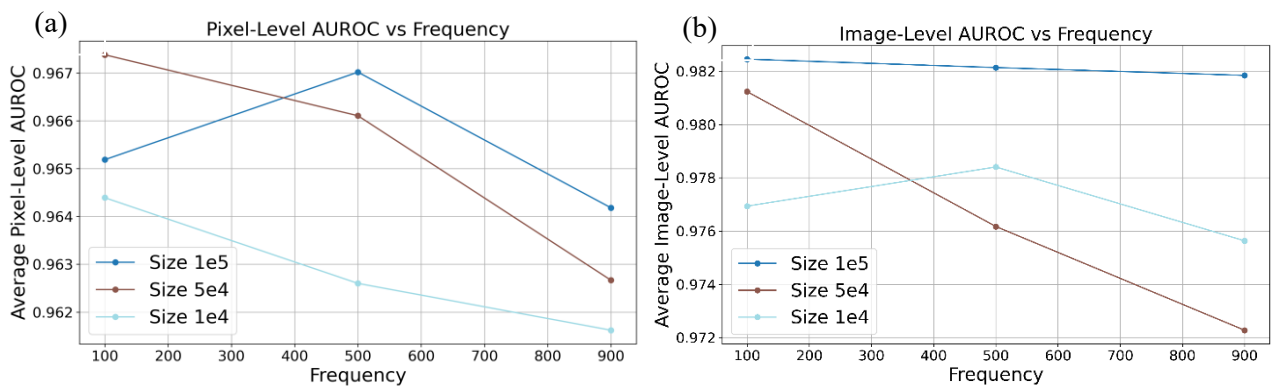

**Figure S3. Effect of subsampling frequency and pool size on model performance on MVTec AD**

(a) Average pixel-level AUROC as a function of the sampling frequency for three subsamples sizes ($1 \times 10^4, 5 \times 10^4, 1 \times 10^5$). (b) Average image-level AUROC under the same conditions. Frequency refers to the interval between subsampling steps (e.g., a frequency of 100 means subsampling every 100 steps).

**Details of Optional Components**

**Prioritized Experience Replay (PER):** In enhancing the DQN, we integrate Prioritized Experience Replay, an advanced technique that improves learning efficiency by modifying the training data sampling strategy.

Unlike conventional experience replay, where transitions are uniformly sampled from a replay buffer, PER prioritizes transitions based on their significance. This prioritization allows the agent to replay important experiences more frequently, accelerating the learning process and improving overall performance. By focusing on transitions with higher expected learning progress, PER helps overcome the inefficiencies of random sampling, enhancing the convergence speed of the DQN.

PER assigns a priority to each experience, making important experiences more likely to be sampled. The priority is typically based on the Temporal-Difference (TD) error of the experience, calculated as:

$$\delta_i = r_i + \gamma \max_{a'} Q(s_{i+1}, a') - Q(s_i, a_i).$$

This prioritization allows the agent to replay important experiences more frequently, accelerating the learning process and improving overall performance. Sampling probability $P(i)$ is determined by:

$$P(i) = \frac{p_i^\alpha}{\sum_k p_k^\alpha},$$

Where $p_i = |\delta_i| + \epsilon$, $\alpha$ controls priority impact, and $\epsilon$ prevents zero priority. Importance-sampling weights $w_i$ correct sampling bias:

$$w_i = \left(\frac{1}{N} * \frac{1}{P(i)}\right)^\beta,$$

Where $N$ is the experience pool size and $\beta$ controls importance sampling degree, typically increasing during training. By focusing on transitions with higher expected learning progress, PER helps overcome the inefficiencies of random sampling, enhancing DQN's convergence speed and optimizing learning.

**Pseudo Anomaly Generation:** To address data imbalance and prevent overfitting, we use the RandAugment-based Pseudo Anomaly Generation (RPAG) strategy from [1] to synthesize anomalies. This approach utilizes a set of image augmentation $T$ which include Flip, Rotate, Transpose, Noise, Distortion, Brightness, Sharpness, Translate and Blur. For each pseudo anomaly, an anomalous image $I^a$ from $D^a$ and a normal image $I^n$ from $D^u$ are selected. A subset of transformations, $\tau \subset T$ is applied to $I^a$ to create $(I^a)' = \tau(I^a)$. The augmented anomalous regions from $(I^a)'$ and then pasted onto the normal image $I^n$, resulting in the pseudo-anomalous image $I_s^a$. If data augmentation is applied, the anomalous patch features from both $I^a$ and its augmented version $I_s^a$ are included in the anomalous subset $S^a$.

**Boundary Study:** Exploration in the RL framework enables flexible and effective feature manipulation. We differentiate normal and abnormal features in $D^a$, denoted as $M^n$ and $M^a$. To improve boundary detection, a subset of abnormal features $M_c^a \subset M^a$ is stored in a memory bank, aiding exploration near-boundary normal features. Given the limited anomalous images and $|M_c^a| = 1000$, the memory overhead is negligible. If the BS strategy is applied, during subsampling, we use the memory bank $M_c^a$ to identify the $K$ closest normal features $S_n' \subset S^n$ and add it to $S^u$. Sampling functions $g_a$ and $g_u$ are executed on $S^a$ and $S^u$, with subsets updated every $T$ epochs.

**Medical Datasets Details**

Brain MRI (Magnetic Resonance Imaging) is a non-invasive imaging technology that produces detailed images of the brain's anatomy, allowing for the detection and diagnosis of various brain conditions.

The IXI dataset includes brain MRI scans from 576 healthy individuals, obtained from several institutions, with a uniform spatial resolution of $0.94 \times 0.94 \times 1.25\ mm^3$, a matrix size of $256 \times 256$, and slice counts varying from 28 to 136. The BraTS 2021 dataset consists of 1,251 subjects diagnosed with Glioblastoma; these images are normalized to a $1 \times 1 \times 1\ mm^3$ resolution, are skull-stripped, and cropped to a consistent size of $240 \times 240 \times 155$.

We adhered to the 3D brain MRI volume preprocessing guidelines found in [2]\cite{luo2023uad3dbrain}, employed the HD-BET tool [3] to skull-strip the T2 volumes from IXI, and aligned all volumes from IXI and BraTS2021 to the Montreal Neurological Institute (MNI) template using the FMRIB Software Library's flirt tool [4,5]. A key difference in our preprocessing compared to SSNet [6] is that, while SSNet simply resamples all images to the standard $1mm$ MNI template without perform co-registration into a common space, we explicitly co-register both the T1- and T2-weighted scans from the IXI and BraTS2021 datasets to a single, consistent template.

**Related Work**

AD is challenging due to the unpredictable, varied, and subtle nature of anomalies, along with the difficulty of collecting and labeling data. The majority of mainstream methods and datasets focus on unsupervised and semi-supervised approaches.

Advanced unsupervised methods leverage deep neural networks to learn complex representations from normal data, identifying deviations from this norm as anomalies. These approaches include reconstruction- and generative-based methods as noted by [7–10]. They generate or reconstruct normal patterns, operating on the premise that they cannot accurately replicate abnormal patterns not encountered during training. This discrepancy between the input and reconstructed output highlights anomalies.

Additionally, synthesizing-based methods, as referenced by [11,12], generate realistic anomalies from normal samples and train models to detect them. Furthermore, embedding-based methods, such as those by [13,14] learning features of normal data and projecting them into an embedding space where abnormal features become more distinguishable. Many advanced methods in this category employ pre-trained models from ImageNet to enhance feature extraction, leveraging the sophisticated training of these large-scale models to boost detection accuracy. The feature extraction protocol using pre-trained models has been shown to contains sufficient and meaningful information for anomaly localization [13,15], we adopt the same protocol to extract the patch representations, which are defined as the state in our DRL algorithm.

Semi-supervised AD leverages a few labeled anomalies to guide the learning of normality distributions. Traditional approaches, like one-class classification, treat anomalies as negative samples [16]. Methods such as Deep SVDD [17], REPEN [18], and MLEP [19] have extended this paradigm using neural networks, distance-based ranking losses, and margin-enhancing techniques. Deep SAD [20] integrates Deep SVDD with entropy minimization to refine representations with labeled data, while further enhancements like multiscale image structures [21] and interpretable anomaly maps [22] have improved detection precision and interpretability. One key insight from these works is that successful algorithms should be grounded in unsupervised methods, with anomalies serving to refine the learned normality distribution. Our algorithm is designed to implicitly adhere to this assumption.

DevNet by [23] employs a deviation loss to drive anomalies away from a prior Gaussian distribution. This is enhanced by CAAD from [24], which adds a confidence prediction module for reliable anomaly scoring. Building on this, [25] integrates DevNet with a top-K multiple-instance learning strategy to provide more generalized representations and utilize gradients for anomaly explanation.

[26] classifies anomalies into seen, pseudo, and latent residuals, synthesizing the latter in latent space to address unknown anomalies. Similarly, [27] uses multi-scale residuals for managing unknown anomalies, while [1] refines normal distribution boundaries using anomalous features. In contrast to these complex approaches, we propose a simpler RL-based classifier that self-explores large environments to effectively localize anomalies and mitigate overfitting.

The integration of DRL in AD is gaining traction, highlighted by a comprehensive review by [28], which discusses RL applications in cybersecurity and streaming data. Further research, including works by [29,30], explores controlled, low-resource, and dynamic methodologies to refine RL in AD. Our model draws inspiration specifically from [31,32]. The former explored AD in non-sequential data within a self-designed environment, while the latter demonstrated the efficacy of deep Q-networks with prioritized experience sampling for multivariate sequential time series. These insights inform our adoption of RL techniques, enhancing our framework to suit image data analysis and leveraging RL's adaptability for dynamic and diverse dataset challenges.

Our research addresses key gaps in surface AD. While traditional DQL-based AD methods often neglect the specific characteristics of image data, our approach integrates representation learning with DRL. This integration not only advances the SOTA in image-based AD but also introduces an approach to establishing boundaries between normal and abnormal data.

**Algorithm Theoretical Analysis**

**Convergence**

In the [33], the convergence of Q-learning is guaranteed if:

1. Learning Rate Conditions: The learning rate $\alpha_t(s, t)$ satisfies:

$$\sum_{t=1}^{\infty} \alpha_t(s, a) = \infty, \quad \sum_{t=1}^{\infty} \alpha_{t^2}(s, a) \leq \infty$$

   which ensures sufficient exploration and diminishing updates

2. Exploration: Every state-action pair (s,a) is visited infinitely often, ensuring all parts of the state space are explored.

3. Finite State and Action Space: The algorithm assumes a finite number of states and actions, which is critical to convergence.

4. Stationary Environment: The underlying Markov Decision Process is assumed to be stationary, meaning the state transition probabilities do not change over time.

Our algorithm keeps using common DQN techniques like RMSprop, $\epsilon$-greedy exploration, and PER, which ensure a diminishing learning rate, effective input space exploration, and prioritized sampling, respectively. Alongside these, we incorporate subsampling and a tailored exploration-exploitation strategy that determines the next state based on the current state. As a result, classical state transitions and strict stationarity do not apply. However, since the data distribution remains static, we treat the state space as effectively stationary, enabling learning from a fixed dataset. Overall, our algorithm satisfies the convergence conditions.

With the exploitation and exploration mechanism, we skip easy and learned points and focus on meaningful points around the decision boundary, which leads to fast convergence.

**Imbalance Analysis**

We've tried a data augmentation technique, commonly used in the AD field, to augment the anomalous samples. While this approach has shown effectiveness on industrial datasets, experiments indicate that the specific augmentation technique we used - RPAG - does not work well for brain MRI data. This is because synthesizing realistic anomalies in brain MRI data poses additional challenges due to its unique characteristics.

Our exploration and exploitation algorithm balances the training samples by sampling normal features with probability $\beta$ and abnormal features with probability $1 - \beta$. Additionally, it carefully selects meaningful features from the normal data, effectively reducing the real size of normality for training. This mechanism ensures a more balanced and representative training process while addressing the inherent data imbalance in anomaly detection tasks.

**Overfitting Analysis**

Introducing stochasticity increases the diversity of training trajectories the model can take, effectively averaging out variability caused by specific training datasets. By incorporating random subsampling and additional randomness in the exploration and exploitation algorithm, the model is encouraged to explore a wider range of states, reducing the likelihood of overfitting to specific regions of the data. Furthermore, the $\epsilon$-greedy algorithm in DQN introduces randomness into the decision-making process, allowing the classifier to make random guesses with probability $\epsilon$, which helps prevent premature convergence to suboptimal decision boundaries.

Theoretically, given a normal feature, our exploration mechanism selects the farthest point if the agent predicts correctly and the closest point if it predicts incorrectly.

1. Correct Prediction (Choosing the Farthest Point)

    When the state $s_t$ is predicted correctly, selecting the farthest point promotes diversity in the training data. Overfitting often results from overemphasis on localized patterns within the training dataset. By targeting the farthest point, the algorithm ensures that the model explores unseen regions of the state space, reducing the risk of overfitting to specific patterns. Additionally, farthest points help expand the decision boundary, encouraging generalization to diverse and unseen inputs. Training on these challenging (farthest) points reduces bias in underrepresented regions, thereby mitigating overfitting to localized patterns.

2. Incorrect Prediction (Choosing the Closest Point)

    When the state $s_t$ is predicted incorrectly, selecting the closest point directs the learning process toward refining local errors. Misclassified points are often located near the decision boundary, making them critical for improving classification accuracy. By focusing on the closest point, the algorithm concentrates updates on these crucial regions, effectively sharpening the decision boundary. Training on nearby points also reduces variance in these regions, resulting in a more stable and reliable boundary. This targeted

approach ensures that the model learns to correct errors while maintaining robustness near the decision boundary.

**One-class Classification Assumption**

As argued by \cite{gornitz2013toward}, successful anomaly detection methods must inherently rely on the unsupervised learning paradigm on normal data due to the presence of unseen outliers. Specifically, labeled anomalous information should be used to refine the boundary of the normal data distribution, rather than establishing a discriminative boundary directly between normal and anomalous data.

Although our agent's policy has two actions (normal vs. anomalous), we do not train it via a cross-entropy loss on both classes. Instead, we

1. Model only the normal distribution by exploring "farthest" and "closest" normal points to expand and refine its one-class boundary, and

2. Use anomalies only as reward signals to nudge that boundary outward—never as positive class targets in a supervised loss.

In this sense, our method is a semi-supervised extension of the one-class assumption: anomalies only refine a model of normality, rather than define a direct discriminative boundary.

**Details of Experiments on Classification Baselines**

Anomalies are rare, unpredictable deviations from normal patterns, making it impractical to construct labeled datasets that capture all possible variations. Their occurrence is often context-dependent and dynamic, differing across domains such as medical imaging or industrial defect detection. This limitation underscores the value of unsupervised and semi-supervised approaches, which learn to detect anomalies by identifying deviations from normal patterns without relying on exhaustive labels. Consequently, most state-of-the-art methods focus on these paradigms, leaving no directly comparable supervised methods for such cases.

In addition, anomalies occupy ≪ 1% of pixels in AD datasets, further less in the used medical data, a severity of imbalance that even weighted losses cannot fully correct and misclassified outliers still dominate gradient updates, causing instability and overfitting.

To demonstrate that our DQN framework is fundamentally different from—and superior to—standard supervised classifiers under extreme class imbalance, we implemented three classification baselines using the identical feature extractor and network architecture as our agent:

1. $C_1$: weighted cross-entropy classifier

2. $C_2$: $C_1$ augmented with RPAG synthetic anomalies

3. $C_3$: $C_2$ plus our subsampling strategy to limit overfitting

Table S3 compares the performance of our method with these three classification strategies using the extracted features. The table demonstrates that the promising performance of our model is primarily attributed to the reinforcement learning framework.

**Table S3. Comparison of classification baselines and our proposed models**

| Dataset | $C_1$ | $C_2$ | $C_3$ | Ours |
|---|---|---|---|---|
| MVTec AD | 74.0/31.7 | 73.0/31.8 | 48.9/58.5 | 99.8/99.3 |
| BTAD | 72.6/36.5 | 70.6/36.0 | 49.0/55.7 | 95.0/98.6 |

I_AUROC/P_AUROC (%) on the MVTec AD and BTAD datasets are reported for three classification strategies – $C_1$: weighted cross-entropy classifier; $C_2$: weighted cross-entropy classifier + RPAG augmentations; $C_3$ weighted cross-entropy classifier + RPAG + subsampling – and for our DQN-All method.

References

1. Yao X, Li R, Zhang J, Sun J, Zhang C. Explicit boundary guided semi-push-pull contrastive learning for supervised anomaly detection. In: *Proceedings of the IEEE/CVF Conference on Computer Vision and Pattern Recognition*. ; 2023:24490-24499.

2. Luo G, Xie W, Gao R, Zheng T, Chen L, Sun H. Unsupervised anomaly detection in brain MRI: Learning abstract distribution from massive healthy brains. *Computers in Biology and Medicine*. 2023;154:106610.

3. Isensee F, Schell M, Pflueger I, et al. Automated brain extraction of multisequence MRI using artificial neural networks. *Human brain mapping*. 2019;40(17):4952-4964.

4. Jenkinson M, Smith S. A global optimisation method for robust affine registration of brain images. *Medical image analysis*. 2001;5(2):143-156.

5. Jenkinson M, Bannister P, Brady M, Smith S. Improved optimization for the robust and accurate linear registration and motion correction of brain images. *Neuroimage*. 2002;17(2):825-841.

6. Zhang Z, Mohsenzadeh Y. Efficient Slice Anomaly Detection Network for 3D Brain MRI Volume. *arXiv preprint arXiv:240815958*. Published online 2024.

7. Guo J, Lu S, Jia L, Zhang W, Li H. Encoder-Decoder Contrast for Unsupervised Anomaly Detection in Medical Images. *IEEE Transactions on Medical Imaging*. Published online 2023:1-1. doi:10.1109/TMI.2023.3327720

8. Kascenas A, Sanchez P, Schrempf P, et al. The role of noise in denoising models for anomaly detection in medical images. *Medical Image Analysis*. 2023;90:102963. doi:10.1016/j.media.2023.102963

9. Schlegl T, Seeböck P, Waldstein SM, Langs G, Schmidt-Erfurth U. f-AnoGAN: Fast unsupervised anomaly detection with generative adversarial networks. *Medical image analysis*. 2019;54:30-44.

10. Kascenas A, Pugeault N, O'Neil AQ. Denoising autoencoders for unsupervised anomaly detection in brain MRI. In: *International Conference on Medical Imaging with Deep Learning*. PMLR; 2022:653-664.

11. Li CL, Sohn K, Yoon J, Pfister T. Cutpaste: Self-supervised learning for anomaly detection and localization. In: *Proceedings of the IEEE/CVF Conference on Computer Vision and Pattern Recognition*. ; 2021:9664-9674.

12. Schlüter HM, Tan J, Hou B, Kainz B. Natural synthetic anomalies for self-supervised anomaly detection and localization. In: *European Conference on Computer Vision*. Springer; 2022:474-489.

13. Liu Z, Zhou Y, Xu Y, Wang Z. Simplenet: A simple network for image anomaly detection and localization. In: *Proceedings of the IEEE/CVF Conference on Computer Vision and Pattern Recognition*. ; 2023:20402-20411.

14. Gudovskiy D, Ishizaka S, Kozuka K. Cflow-ad: Real-time unsupervised anomaly detection with

localization via conditional normalizing flows. In: *Proceedings of the IEEE/CVF Winter Conference on Applications of Computer Vision.* ; 2022:98-107.

15. Roth K, Pemula L, Zepeda J, Schölkopf B, Brox T, Gehler P. Towards total recall in industrial anomaly detection. In: *Proceedings of the IEEE/CVF Conference on Computer Vision and Pattern Recognition.* ; 2022:14318-14328.

16. Görnitz N, Kloft M, Rieck K, Brefeld U. Toward supervised anomaly detection. *Journal of Artificial Intelligence Research*. 2013;46:235-262.

17. Ruff L, Vandermeulen R, Goernitz N, et al. Deep one-class classification. In: *International Conference on Machine Learning*. PMLR; 2018:4393-4402.

18. Pang G, Cao L, Chen L, Liu H. Learning representations of ultrahigh-dimensional data for random distance-based outlier detection. In: *Proceedings of the 24th ACM SIGKDD International Conference on Knowledge Discovery & Data Mining.* ; 2018:2041-2050.

19. Liu W, Luo W, Li Z, Zhao P, Gao S, others. Margin Learning Embedded Prediction for Video Anomaly Detection with A Few Anomalies. In: *IJCAI*. Vol 3. ; 2019:023-3.

20. Ruff L, Vandermeulen RA, Görnitz N, et al. Deep semi-supervised anomaly detection. *arXiv preprint arXiv:190602694*. Published online 2019.

21. Ruff L, Vandermeulen RA, Franks BJ, Müller KR, Kloft M. Rethinking assumptions in deep anomaly detection. *arXiv preprint arXiv:200600339*. Published online 2020.

22. Liznerski P, Ruff L, Vandermeulen RA, Franks BJ, Kloft M, Müller KR. Explainable deep one-class classification. *arXiv preprint arXiv:200701760*. Published online 2020.

23. Pang G, Shen C, Van Den Hengel A. Deep anomaly detection with deviation networks. In: *Proceedings of the 25th ACM SIGKDD International Conference on Knowledge Discovery & Data Mining.* ; 2019:353-362.

24. Zhang J, Xie Y, Pang G, et al. Viral pneumonia screening on chest X-rays using confidence-aware anomaly detection. *IEEE transactions on medical imaging*. 2020;40(3):879-890.

25. Pang G, Ding C, Shen C, Hengel A van den. Explainable deep few-shot anomaly detection with deviation networks. *arXiv preprint arXiv:210800462*. Published online 2021.

26. Ding C, Pang G, Shen C. Catching both gray and black swans: Open-set supervised anomaly detection. In: *Proceedings of the IEEE/CVF Conference on Computer Vision and Pattern Recognition.* ; 2022:7388-7398.

27. Zhang H, Wu Z, Wang Z, Chen Z, Jiang YG. Prototypical residual networks for anomaly detection and localization. In: *Proceedings of the IEEE/CVF Conference on Computer Vision and Pattern Recognition.* ; 2023:16281-16291.

28. Arshad K, Ali RF, Muneer A, et al. Deep reinforcement learning for anomaly detection: A systematic review. *IEEE Access*. 2022;10:124017-124035.

29. Zhong C, Gursoy MC, Velipasalar S. Controlled sensing and anomaly detection via soft actor-critic reinforcement learning. In: *ICASSP 2022-2022 IEEE International Conference on Acoustics, Speech and Signal Processing (ICASSP)*. IEEE; 2022:4198-4202.

30. Li Y, Wu J. Low latency cyberattack detection in smart grids with deep reinforcement learning. *International Journal of Electrical Power & Energy Systems*. 2022;142:108265.

31. Fährmann D, Jorek N, Damer N, Kirchbuchner F, Kuijper A. Double deep q-learning with prioritized experience replay for anomaly detection in smart environments. *IEEE Access*. 2022;10:60836-60848.

32. Pang G, van den Hengel A, Shen C, Cao L. Toward deep supervised anomaly detection: Reinforcement learning from partially labeled anomaly data. In: *Proceedings of the 27th ACM SIGKDD Conference on Knowledge Discovery & Data Mining*. ; 2021:1298-1308.

33. Watkins CJ, Dayan P. Q-learning. *Machine learning*. 1992;8:279-292.